# A Self-Supervised Denoising Strategy for Underwater Acoustic Camera Imageries

Xiaoteng Zhou, Katsunori Mizuno, Yilong Zhang

*Graduate School of Frontier Sciences, The University of Tokyo, Kashiwa, Chiba, 277-8563, Japan*

*Abstract*—In low-visibility marine environments characterized by turbidity and darkness, acoustic cameras serve as visual sensors capable of generating high-resolution 2D sonar images. However, acoustic camera images are interfered with by complex noise and are difficult to be directly ingested by downstream visual algorithms. This paper introduces a novel strategy for denoising acoustic camera images using deep learning techniques, which comprises two principal components: a self-supervised denoising framework and a fine feature-guided block. Additionally, the study explores the relationship between the level of image denoising and the improvement in feature-matching performance. Experimental results show that the proposed denoising strategy can effectively filter acoustic camera images without prior knowledge of the noise model. The denoising process is nearly end-to-end without complex parameter tuning and post-processing. It successfully removes noise while preserving fine feature details, thereby enhancing the performance of local feature matching.

*Keywords—low-visibility marine environments, acoustic camera, image denoising, deep learning, local feature matching, sonar*

## I. INTRODUCTION

In recent years, to meet the demands of sustainable human development, coastal cities have seen increasingly dense and large-scale infrastructure construction, extending into deeper sea areas. In this process, scientific and effective management measures must be implemented to ensure the sustainability of the marine environment and the health of the ecosystem. It is particularly important to emphasize that regular inspection and maintenance of marine structures, such as ports, oil platforms, and offshore wind farms, are critical measures to ensure their long-term stable operation and to minimize their impact on marine ecosystems. Only through scientific inspection and maintenance can potential issues be promptly identified and rectified, preventing environmental pollution and ecological damage, thereby promoting sustainable marine development.

To achieve this goal, effectively perceiving low-visibility marine environments is the first step. In shallow water areas, this task is typically performed by divers using underwater optical cameras. However, with the deepening of marine structures, more complex underwater environments (including turbidity, darkness, and hazards) prevent divers from reaching the investigation sites and render optical cameras ineffective. Therefore, underwater perception requires more stable sensing technology to support it. One notable advancement in this field is the emergence of acoustic cameras, a specialized subset of 2D forward-looking sonar (FLS). Prominent examples of acoustic cameras include dual-frequency identification sonar (DIDSON) and adaptive resolution imaging sonar (ARIS). These sensors can generate high-resolution sonar images that are very similar to optical imaging, thereby enhancing human understanding of underwater conditions [1]. Moreover, acoustic cameras are flexible and can be easily installed on underwater robots, such as Autonomous Underwater Vehicles (AUVs) or Remotely Operated Vehicles (ROVs), to replace divers performing underwater perception operations, as depicted in Fig. 1.

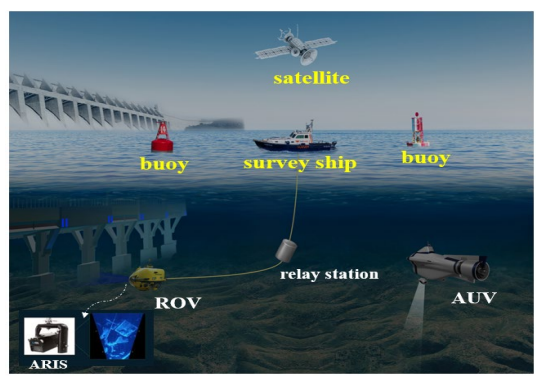

Fig. 1. Demonstration of underwater perception using acoustic cameras.

Regrettably, basic image processing algorithms for acoustic cameras have significantly lagged behind advances in hardware. For instance, image denoising, which is typically the first step in almost all image-based visual applications, remains a major challenge for acoustic images. Currently, there is no dedicated denoising algorithm specifically designed for acoustic images. Due to the nonlinearly superimposed and complex nature of the noise in acoustic images, it becomes challenging to acquire prior knowledge of the noise model, which is the initial step in a handcrafted filter. With the advent of deep learning, there are now opportunities to overcome these limitations by adopting data-driven approaches instead of handcrafted filters.

## II. RELATED RESEARCH

### A. Acoustic Camera Images Noise Analysis

The real marine environments are complex, and due to the presence of numerous scatterers and rough interfaces that cause strong reverberation, there is a significant amount of background noise in acoustic images. The main types of noise are as follows: (i) Point clutter: this is a discrete noise caused by scattering objects, bubbles, and interferences. (ii) Acoustic noise: this is an unexpected signal originating from the external environment, typically including water flow noise, mechanical noise, and marine biological noise.(iii) Reflection noise: it refers to the noise that occurs when sound waves reflect at interfaces between objects.(iv) Instrument noise: this is introduced by the inherent components of the sonar sensor itself and the signal processing, including electronic noise, amplifier noise, and converter noise.

### B. Analysis of the Influence of Noise on Feature Matching

Local feature matching is the basis for the visual application. For acoustic image feature matching, denoising is a crucial step. This is because noise directly affects the core steps of local feature matching: feature detection and feature description.

*1)* Feature detection: noise will introduce additional details or blur, making it difficult to accurately detect the positions and intensities of feature points. This may result in detected feature points with low repeatability, insufficient numbers, and wrong positions. However, feature detection is a critical prerequisite and the first step in local feature matching, which directly affects the final accuracy.

*2)* Feature description: since noise changes the local intensity of the image, first of all, it will directly lead to deviations in finding key descriptor information around feature points. This condition ultimately results in the construction of complex and blurred feature descriptor structures, which subsequently reduces the similarity between corresponding descriptors. Secondly, feature points at different positions may construct similar descriptors due to noise. Thirdly, noise could make feature descriptors more sensitive to interference, thereby reducing their robustness in complex environments.

Fig. 2 offers an illustrative example showcasing the influence of noise on acoustic image feature matching. It can be seen that after denoising the image pairs, the accuracy of the feature-matching results can be significantly improved.

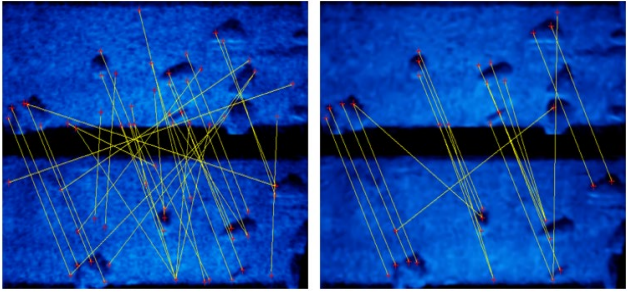

Fig. 2. Matching results of raw images (left) and after denoising (right).

### C. Related Acoustic Image Denoising Research

The literature survey reveals that the main solution in the field of acoustic camera image denoising is to assume a specific noise model on the image and then apply classical image denoising filters for processing. Although some studies have optimized these denoising filters, they still have obvious limitations, such as only being able to work in fixed detection scenes. In addition, these methods highly rely on prior knowledge of the noise model to adjust the parameters of the filter, which easily leads to deep denoising or weak denoising effects. Few researchers have analyzed the impact of acoustic image denoising on downstream visual tasks. For instance, does deep denoising enhance the accuracy of downstream tasks? Determining the optimal balance between effective denoising and the preservation of features remains a significant challenge.

Some common denoising approaches are shown in Table I.

TABLE I. SOME ACOUSTIC CAMERA IMAGE DENOISING APPROACHES

|   | Sensor type | Denoising methods | Source |
|---|---|---|---|
| 1 | DIDSON | Gaussian filter | [2] |
| 2 | DIDSON | Averaging multiple images | [3] |
| 3 | ARIS | Wavelet denoising (db2) | [4] |
| 4 | DIDSON | Anisotropic denoising | [5] |

### D. Motivation and Contribution

Based on the literature survey, it is clear that the noise present in DIDSON and ARIS acoustic images is of a complex nature, and acquiring prior knowledge about this noise proves to be challenging. Currently, the dominant denoising techniques employed for these images are adapted from the field of optical image denoising. Recently, deep learning for self-supervised denoising from only a single noisy image has been developed. These models are not strict about the number of images used for training, which solves the limitation of insufficient acoustic image samples. Among the denoising models with outstanding performances are NBR2NBR [6] and Blind2UnB [7]. Considering the inherent properties and imaging geometry of underwater acoustic images, the method design of this paper is inspired by [6], and the following main contributions are made.

*1)* A deep learning-based self-supervised denoising method designed for underwater acoustic camera images is proposed. This framework can effectively denoise acoustic images without making any assumptions about the scene or requiring prior knowledge of the noise characteristics.

*2)* The denoising framework is directly tailored for feature-matching tasks, revealing the potential impact of denoising on acoustic image feature matching. It also provides interfaces to support downstream tasks based on local feature matching, such as image mosaicking, 3D reconstruction, and SLAM.

*3)* The denoising approach requires minimal parameter tuning and can be used to denoise single or multiple images in a near end-to-end manner, regardless of the imaging geometry.

*4)* Compared to classical denoising methods in the field of acoustic image processing, the proposed method demonstrates superior denoising performance, particularly on practical acoustic camera images with complex noise patterns.

## III. RESEARCH METHODOLOGY

The precise impact of acoustic camera image denoising on subsequent feature-matching processes remains ambiguous. deep denoising may induce over-smoothing of acoustic images, potentially resulting in the loss of repeatable feature points. Conversely, weak denoising might fail to sufficiently remove noise, leading to the generation of biased descriptors. To address these issues, the proposed denoising framework is designed to balance these trade-offs, as depicted in Fig. 3.

The experimental dataset comprises approximately 8000 acoustic camera images sourced from the image gallery on Sound Metrics [8]. These images have undergone preprocessing to enhance their suitability for analysis. This study maintained the original fan-shaped shape of the acoustic camera image.

A ratio $M$ ($M \in [0.5,1]$) of the dataset is used for training the self-supervised denoising model (training dataset), while the remaining dataset (test dataset) is used for testing the denoising model. In addition, the dataset used for training is randomly transformed by translation to generate a dataset for testing the feature-matching performance (matching dataset), which is used to analyze the impact of denoising on local feature matching.

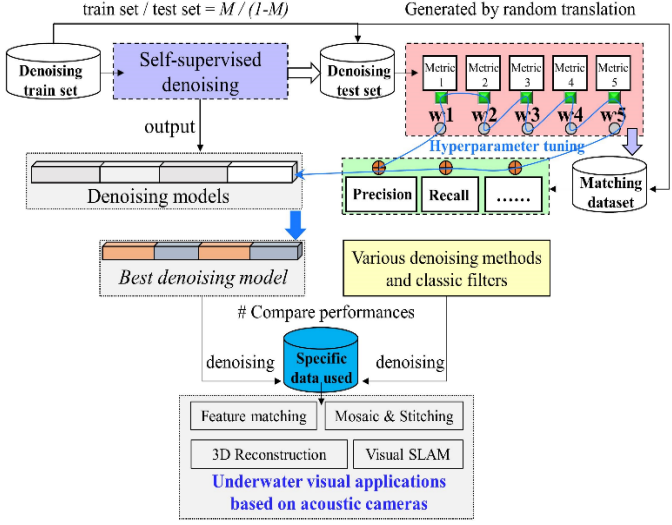

Fig. 3. The workflow of the analysis framework proposed in this study.

As illustrated in Fig. 3, to assess the image denoising effects, five metrics are introduced: PSNR (Peak Signal-to-Noise Ratio), SSIM (Structural Similarity), EPI (Edge Preservation Index) [9], TV (Total Variation) [10], and BRISQUE (Blind/Referenceless Image Spatial Quality Evaluator) [11]. PSNR, SSIM, and EPI are reference indexes, and TV and BRISQUE are non-reference indexes. Precision, Recall and Time indicators are used to evaluate the feature-matching performances after denoising.

The five evaluation metrics introduced are assigned weights ($W_i, \sum_{j=0}^{i} W_j = 1$) and the best combination is obtained via genetic algorithm hyperparameter tuning to achieve the best matching performance score ($Score = \sum (\text{RANK}(i) * W_i)$).

Finally, the best denoising model is then selected by the best combination. Then use the best model to process the practical acoustic camera images (specific data) acquired in our previous experiment [12] to verify the denoising framework.

Fig. 4 shows the training pipeline for self-supervised denoising of acoustic images. The body section is the complete view of the training. Generate a pair of images from noisy acoustic images using adjacent subsamplers. The denoising network model uses the subsampled images as input and target. The loss consists of two parts: the upper part calculates the $Loss1$ between the network output and the noise target; the lower part, the $Loss2$ is further added considering the difference between the subsampled noisy image and the ground-truth value. It should be mentioned that neighborhood subsamplers (block in yellow) appearing twice are the same. The bottom right part is an inference demo using the trained acoustic denoising model.

The loss function for network training is defined as follows:

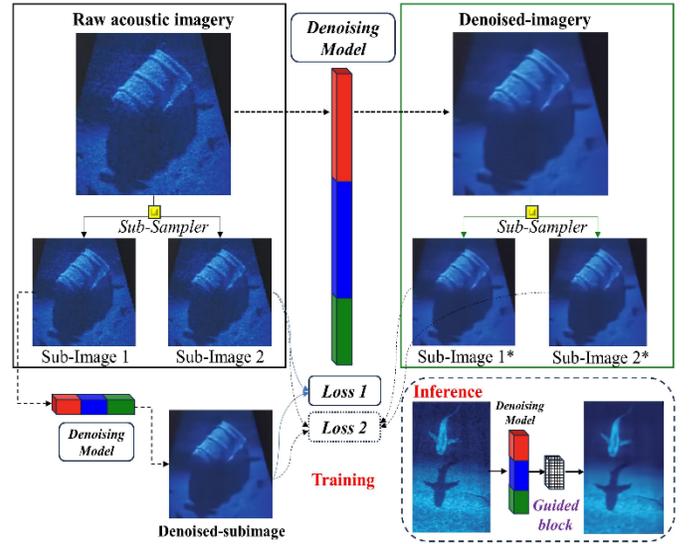

Fig. 4. The pipeline of self-supervised denoising strategy on acoustic images.

$$Loss = Loss1 + \gamma \cdot Loss2 \quad (1)$$

$$Loss1 = \|Denoise(SubImage1) - SubImage2\|_2^2 \quad (2)$$

$$Sub* = (SubImage1* - SubImage2*) \quad (3)$$

$$Loss2 = \|Denoise(SubImage1) - SubImage2 - Sub*\|_2^2 \quad (4)$$

where $Denoise(\theta)$ is the denoising function, for a raw noisy acoustic image, two sub-samples $SubImage1$ and $SubImage2$ are taken. The $Loss1$ is computed between the denoised image $Denoise(SubImage1)$ obtained using the denoising network and $SubImage2$, which represents the target noise. It could calculate the reconstruction error between the network output and the noise target. And, a regularization term is introduced in $Loss2$ to alleviate the over-smoothing of the output image caused by the sampling method. $\gamma$ is a hyperparameter that controls the denoising strength.

The training of the self-denoising model is achieved by acquiring noise pairs through the sub-pixel sampling of the noisy images. This denoising strategy is flexible, requires minimal training samples, and exhibits good generalization. However, this strategy also has notable limitations. For example, approximation by neighboring pixel sampling methods still leads to over-smoothing of the denoised image, while subsampling will destroy the continuity of the detail structure. To address this issue, this paper introduces a guided filtering [13] block to improve these shortcomings.

Specifically, this study designed a fine feature-guided block to enhance the denoising model output (first-stage denoised) through guided filtering and visual saliency detection. As shown in Fig. 5, assuming $p$ represents the input noisy image, which is the raw acoustic camera image without any processing, an $I$ represents the guide image, which is the first-stage denoised image obtained through the self-supervised denoising strategy. Then, the fine feature-guided block is used to guide the transformation of the first-stage denoised image to obtain the final denoised image $q$. At this point, $q$ is nearly free from noise interference while recovering structural features, such as edges and corners. In this process, the filtering processing for one pixel of an image can be expressed as follows.

$$q_i = \sum_j W_{ij}(I) p_j \quad (5)$$

where $q_i$ denotes the denoised pixel value at position $i$, $p_j$ represents the pixel value of the input raw image at position $j$, and $W_{ij}$ denotes the weight assigned to the pixel value $p_j$ based on its relationship with the guided image $I$.

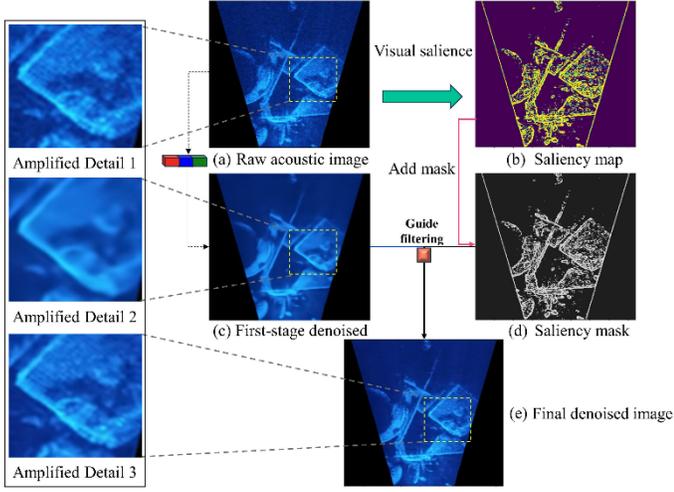

Fig. 5. Schematic diagram of the feature-guided block on acoustic images.

**Algorithm 1** Fine-features guided block for acoustic images

1: **Input**: An acoustic image after first-stage denoising.

2: *STEP 1*: Convert the image to the LAB color space.

3: *STEP 2*: Extract the L channel (luminance).

4: *STEP 3*: Calculate the gradients of the L channel in the x and y directions using the Sobel operator.

5: *STEP 4*: Calculate the square root of the gradient and normalize it.

6: *STEP 5*: Obtain the saliency map from the normalized gradient magnitude image by Otsu algorithm.

7: *STEP 6*: Generate the mask of the saliency map.

8: *STEP 7*: Implement guided filtering on the mask map.

9: **Output**: Final denoised acoustic image

Fig. 5 and Algorithm 1 explain the workflow of the feature-guided block in detail. The yellow line area is used to compare the feature preservation effects.

IV. EXPERIMENT AND VERIFICATION

In the first part of the experiment, a random selection of images was taken from the training dataset for denoising testing. The acoustic image targets encompassed barrel, propeller, wreckage, marine organisms, and. The images were randomly selected, and the noise types present in the images were not identified. The performances of the self-supervised denoising model are shown in Fig. 6.

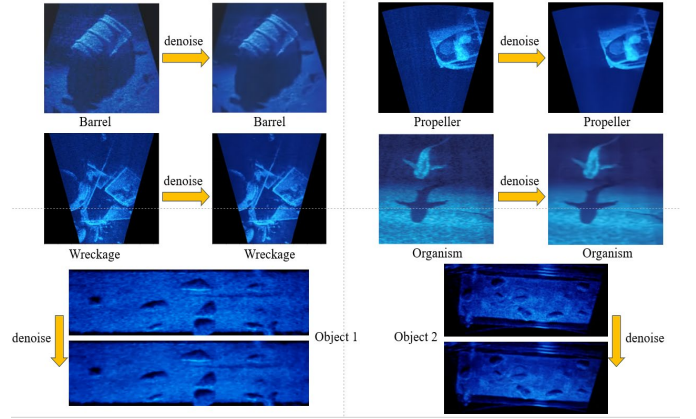

Fig. 6. The denoised effects of our proposal on images of various objects.

Based on the comparison results depicted in Fig. 6, it is evident that the self-supervised denoising approach proposed in this study is capable of effectively removing complex noise from underwater acoustic camera images. Notably, this method does not rely on prior knowledge of the noise model or require tedious parameter tuning. Its flexibility enables it to overcome the challenge imposed by a limited number of training datasets, and it provides valuable preprocessing capabilities for machine vision tasks that rely on underwater acoustic camera images.

In the second part of the experiment, this paper evaluates the proposed denoising approach against several classical denoising techniques as well as recent state-of-the-art learning-based denoising models. The comparison results are illustrated in Fig. 7, where red and yellow areas are highlighted to indicate the extent of feature details preservation.

Based on the comparative analysis of denoising results, better denoising effects are observed in the Anisotropic method, Blind2UnB model, NBR2NBR model, and our proposal. However, Wavelet denoising, Gaussian filter, and Median filter were found to have limited efficacy in effectively eliminating noise, which can be attributed to the complex superposition of noise on acoustic images. The reliance on handcrafted filters, which lack the ability to comprehensively handle diverse noise types, may have contributed to this limitation. In contrast, our proposal demonstrates superior denoising results compared to the Blind2UnB model and NBR2NBR model, while preserving the feature details of the images. These details provide valuable information regarding the material composition of the target and offer critical insights for downstream vision applications. Acoustic camera images inherently face challenges such as low resolution, high signal-to-noise ratio (SNR), and pronounced distortion, which result in the loss of important information. Therefore, it is unfavorable to further compromise the fine features during the denoising phase. The denoising method presented in this paper achieves a better balance between noise reduction and fine feature preservation, providing an effective solution to address these preprocessing challenges.

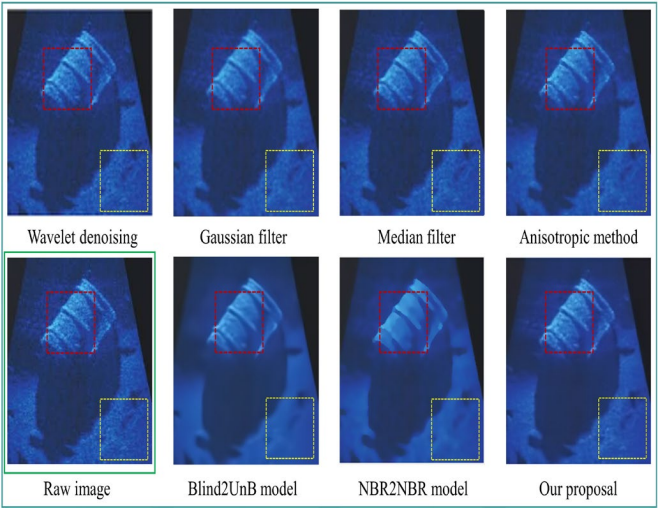

Fig. 7. Comparison of results after applying different denoising methods.

In this study, the Power Spectral Density (PSD) was introduced as a measure of the energy distribution in the frequency domain for assessing the performances of denoising approaches. The raw noisy image and the images denoised by the Anisotropic method, Blind2UnB_model, and our proposal are displayed. The comparative results are illustrated in Fig. 8.

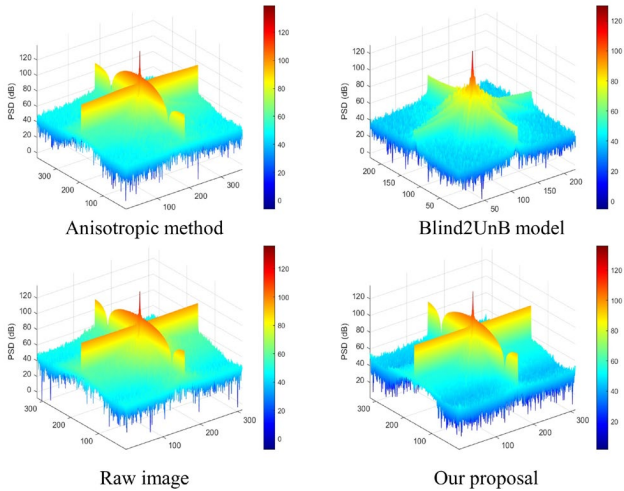

Fig. 8. The PSD maps of the denoised images using different methods.

An analysis of Fig. 8 indicates that the noise observed in the acoustic images predominantly manifests as mid-frequency energy within the transitional regions. High-frequency energy typically corresponds to rapidly changing details and edges, while low-frequency energy corresponds to dull backgrounds without distinct features. When compared to the energy distribution of the original acoustic image, it is evident that the Anisotropic method exhibits limited noise removal, as indicated by the substantial presence of mid-frequency energy. Although the learning-based competitive model Blind2UnB can achieve an obvious noise reduction, it does so at the expense of a substantial amount of local feature details. In contrast, our proposed method not only effectively mitigates noise interference but also preserves the integrity of local features.

In the third part of the experiment, this study investigates the potential mapping relationship between the evaluation of feature-matching performance and denoising performance from a big data perspective. Specifically, 100 denoising models are randomly trained with the proposed self-supervised denoising framework. Subsequently, these 100 models are employed to denoise the training set of acoustic camera images, resulting in a denoised training set that is used to generate a matching dataset, enabling the completion of local feature matching.

This study used the classic SIFT algorithm [14] to achieve local feature matching. The algorithm was implemented based on the OpenCV library [15] in Python, and the internal parameters were all default values. The proposed framework reserves an interface for feature matching, which can test more classical matching algorithms as well as advanced learning-based matching algorithms. The two processes of feature detection and description can also be evaluated independently.

Various metrics, including PSNR, SSIM, EPI, TV, and BRISQUE, are employed as evaluation indicators for denoising performance. It is important to note that these metrics are not fixed, as our framework offers an interface to incorporate various denoising evaluation metrics, encompassing both manually designed and learning-based metrics, for further examination. Moreover, recall and precision are selected as evaluation metrics for feature-matching performances.

*(1) Recall*: It is a metric used to measure the ability of a method to detect and match correct feature points. It reflects the proportion of correctly matched feature points among all relevant keypoints. A high value of recall indicates that the algorithm is able to effectively capture most of the true matching keypoints, thereby providing more comprehensive and accurate feature matching results.

$$Recall = \frac{\text{num}(correct\ matches)}{\max(keypoints\ 1, keypoints\ 2)} \quad (6)$$

*(2) Precision*: It represents the proportion of truly correct matches among all the putative matches. A higher precision value indicates that the system is able to accurately match feature points to the correct locations, reducing the incorrect matches.

$$Precision = \frac{\text{num}(correct\ matches)}{\text{num}(putative\ matches)} \quad (7)$$

Finally, the interactions between different metrics are explored by computing the correlation coefficient matrix and visualizing it through heatmaps.

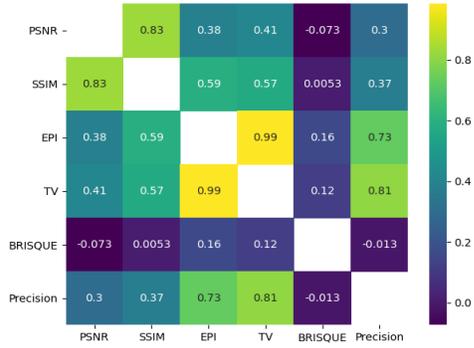

Fig. 9. Correlation analysis results between precision and various denoising evaluation metrics.

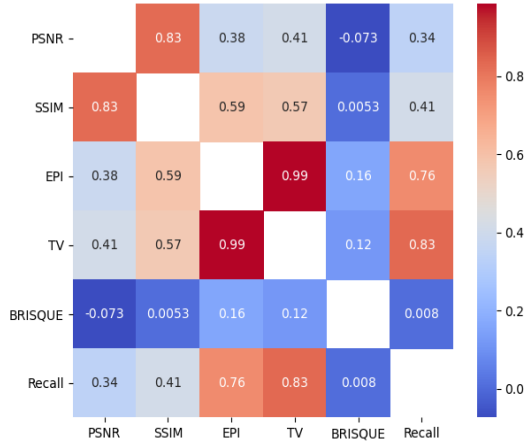

Fig. 10. Correlation analysis results between recall and various denoising evaluation metrics.

From the results in Fig. 9 and 10, it could be observed that the correlation coefficients between Precision and PSNR/SSIM are 0.3 and 0.37, respectively, while the correlation coefficients between Recall and PSNR/SSIM are 0.34 and 0.41, respectively. This suggests a modest positive correlation between PSNR/SSIM and Precision/Recall when assessing image quality. Nonetheless, the relatively low correlation coefficients indicate that the strength of this association may be limited. On the other hand, the correlation coefficients between Precision and EPI/TV are 0.73 and 0.81, respectively, while the correlation coefficients between Recall and EPI/TV are 0.76 and 0.83, respectively, showing a stronger positive correlation. This indicates that the information reflected by the EPI/TV metrics in image denoising is more closely associated with the improvement in feature-matching performance, with higher EPI/TV values often associated with higher feature-matching performance. However, compared to other metrics, the correlation coefficient between BRISQUE and Precision/Recall is below 0.05. This suggests that there is almost no evident linear relationship between BRISQUE and the metrics for evaluating downstream feature-matching performance.

These interesting findings provide valuable insights for the design and improvement of future image-denoising algorithms. It should be noted that in this study, we only explored the linear relationship between denoising evaluation metrics and feature-matching evaluation metrics through correlation coefficients. In future research, we need to consider more factors and methods to investigate the nonlinear relationships among these metrics.

In the fourth part of the experiment, this study verifies the performances of the proposed denoising approach on our previous experiment [12]. This dataset used for validation is a concrete plate covered with particles detected by an ARIS acoustic camera. This dataset has the characteristics of uniform target intensity, clear edges, and clear features, which are very representative. To highlight the effect of image denoising and compare the results of feature matching, we cropped the regions of interest (ROIs). The testing dataset is captured from the raw acoustic file according to the frame rate, a total of 10 pictures, divided into 5 image pairs for denoising and feature matching, and the resolution of each image is 600 x 190.

Experimental scene and data samples are shown in Fig. 11.

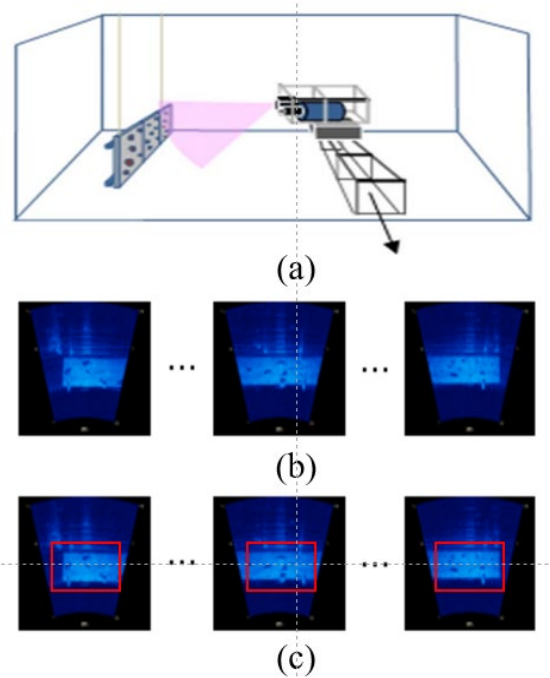

Fig. 11. Practical acoustic camera images acquisition scenarios and samples.

The whole denoising framework was validated through a combination of qualitative and quantitative evaluations, which encompassed both denoising and feature-matching tasks. The denoising quality was evaluated using PSNR and SSIM to assess the denoising performances. The metrics TV and EPI were used to evaluate the edge preservation capability. Additionally, the image quality assessment metric BRISQUE was utilized to evaluate the quality of the denoised images. Table II presents the average evaluation results of various methods on the denoised acoustic images. Fig. 12 displays the denoising results on one acoustic image sample through various methods. Fig. 13 and Fig. 14 illustrate the enhancement in feature-matching performance achieved through denoising.

TABLE II.  DENOISING EVALUATION RESULTS ON ACOUSTIC DATASET

| Denoising methods | Evaluation metrics | | | | |
|---|---|---|---|---|---|
| | PSNR | SSIM | EPI | TV | BRISQUE |
| Raw_images | | | | 7290 | 52.04 |
| Mean_filter | 42.71 | 0.9862 | 0.8229 | 6079 | 52.14 |
| Median_filter | 38.11 | 0.9513 | 0.6815 | 5110 | 61.38 |
| Gaussian_filter [2] | 41.35 | 0.9816 | 0.7880 | 5836 | 52.51 |
| Bilateral_filter | 52.64 | 0.9979 | 0.9382 | 6887 | 56.92 |
| Wavelet_denoising [4] | 48.11 | 0.9947 | 0.9424 | 7006 | 56.33 |
| Anisotropic_method [5] | 35.99 | 0.9253 | 0.5961 | 4531 | 57.40 |
| NBR2NBR_model [6] | 33.79 | 0.8967 | 0.5819 | 4541 | 71.18 |
| Blind2UnB_model [7] | 32.40 | 0.8181 | 0.5219 | 4279 | 62.0 |
| Our proposal | 38.23 | 0.9168 | 0.7507 | 5841 | 58.6 |

To evaluate the impact of various denoising methods on the acoustic camera image feature matching, this part uses *Recall* and *Precision* as quantitative evaluation. In addition, this part also conducted a visual assessment of the matching results after denoising for qualitative evaluation. This combined approach comprehensively analyzes the impact of various denoising methods on local feature matching.

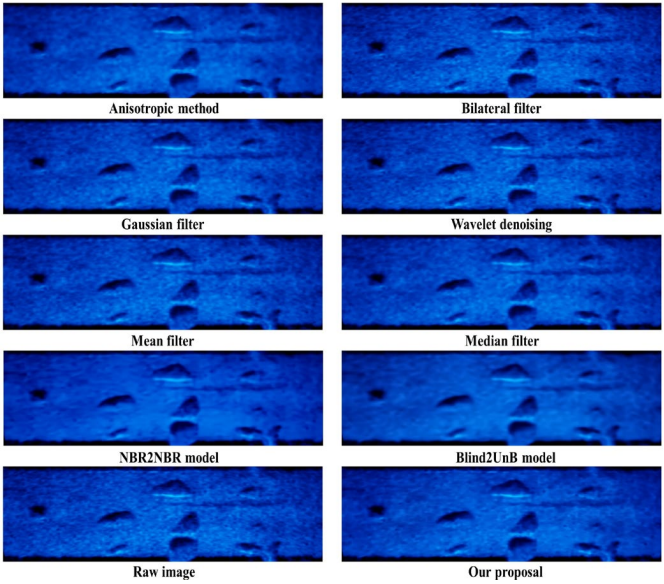

Fig. 12. Comparison of the denoising effect of one acoustic image sample.

The findings indicate that our proposed method efficiently eliminates complex noise while simultaneously maximizing the retention of fine details, thereby enhancing the performance of feature-matching algorithms. Additionally, it could be found that learning-based denoising models via data-driven are more effective than using the most handcrafted filters on the acoustic camera images.

Considering the assessment outcomes delineated in Table II, it becomes apparent that prevalent image denoising metrics, such as PSNR and SSIM, inadequately capture the impact of denoising on local feature matching. It should be noted that if the noise model does not match the actual noise in the image, the algorithm may not be able to effectively remove the noise, which may cause the PSNR and SSIM values to be artificially high, such as when using a mean filter.

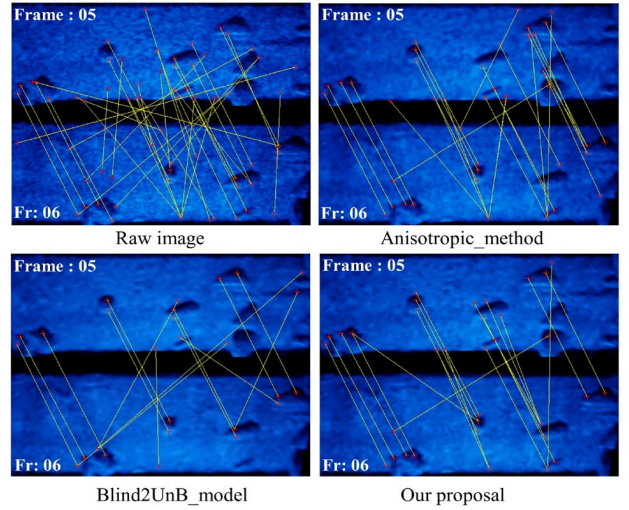

Fig. 13. Frame 5-6 matching results after denoising (Precision top 3 demos).

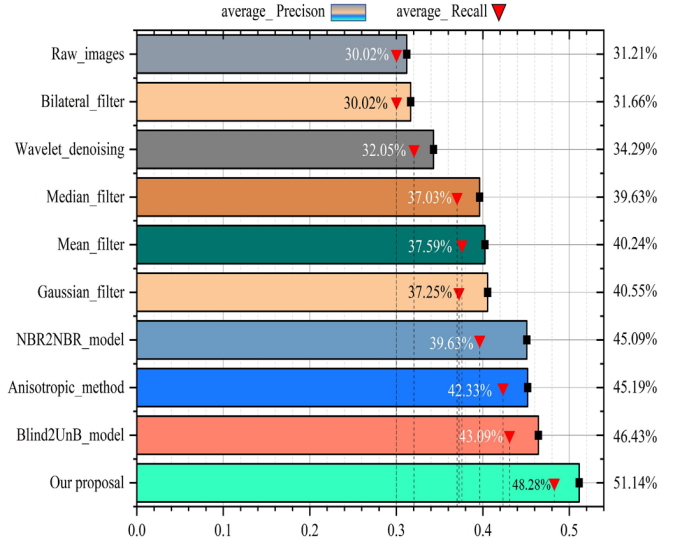

Fig. 14. Feature matching metrics for acoustic image dataset after denoising.

Based on the results in Fig. 14, it can be inferred that image-denoising preprocessing enhances feature-matching performance on acoustic images. However, the appropriate level of denoising should be determined on a case-by-case basis. For acoustic images with low resolution and low SNR, extensive denoising is not always the best option. In other words, there is a coupling relationship between denoising levels and downstream feature-matching performance, where higher denoising metrics do not necessarily ensure improved feature-matching performance.

Deep learning-based denoising approaches significantly enhance matching performance. Furthermore, as evidenced by the data analysis, the EPI and TV indicators highlight the importance of edge retention in feature matching. Therefore, for practical underwater applications using acoustic camera images, selecting suitable denoising levels based on specific requirements is essential.

## V. DISCUSSION

This letter conducts an in-depth study on the denoising of acoustic camera images. Given the fundamental differences between acoustic imaging mechanisms and optical imaging mechanisms, the feature distributions and patterns presented in acoustic camera images exhibit significant disparities compared to optical images. Consequently, the manual design of denoising filters in a theory-driven manner poses substantial challenges. To address this challenge, this study explores data-driven deep learning denoising strategies and selects algorithms suitable for irregular image geometries characteristic (fan or wedge) of acoustic camera images, demonstrating significant efficacy.

Although this research is still in its preliminary stages, we believe that there is further potential for improvement in the denoising performances of the model through strategies such as hyperparameter optimization, expansion of training datasets, and domain transfer in images. The achievements of this paper not only offer innovative approaches to denoising tasks in acoustic camera images but also provide valuable references for denoising research in other types of sonar images. We hope that this study will serve as a starting point in this field, inspiring more research on denoising acoustic camera images and other sonar images, and contributing to the advancement of future acoustic sensing technology.

## VI. CONCLUSIONS

This study proposed a self-supervised deep denoising framework specifically applied to underwater acoustic camera images and evaluated it by validating it on real acoustic datasets. The experimental results indicate that our approach effectively removes noise from the image while retaining fine feature details, without requiring any prior assumptions about the noise model or complex post-processing. In addition, the correlation between existing image-denoising evaluation systems and feature-matching tasks is revealed through hyperparameter tuning. Our proposed method exhibits the most notable enhancement in local feature matching when compared to alternative denoising techniques. This discovery serves as a pivotal benchmark for advancing research in the domain of feature matching utilizing acoustic camera imageries.

In future research, we will explore the impact of target materials on the performance of various denoising algorithms in acoustic images, as the properties of the target materials directly affect the acoustic imaging results. Additionally, we plan to integrate the training of denoising models with the training of downstream visual task models (such as image identification and segmentation) to form a comprehensive acoustic image processing architecture. This joint training approach aims to optimize the performance of the entire sonar image processing pipeline, thereby enhancing the practicality and accuracy of acoustic camera images in various marine application scenarios.